\documentclass[letterpaper]{article} 
\usepackage{aaai2026}  
\usepackage{times}  
\usepackage{helvet}  
\usepackage{courier}  
\usepackage[hyphens]{url}  
\usepackage{graphicx} 
\usepackage{natbib}  
\usepackage{caption} 
\usepackage{algorithm}
\usepackage{algpseudocode}
\usepackage{booktabs}
\usepackage{bm}
\usepackage{amsfonts} 
\usepackage{amsmath}
\usepackage[utf8]{inputenc} 
\usepackage{subcaption}
\usepackage{xcolor}
\usepackage{colortbl}

\usepackage{newfloat}
\usepackage{listings}
\DeclareCaptionStyle{ruled}{labelfont=normalfont,labelsep=colon,strut=off} 
\floatstyle{ruled}
\newfloat{listing}{tb}{lst}{}
\floatname{listing}{Listing}
\title{HumanGenesis: Agent-Based Geometric and Generative Modeling for Synthetic Human Dynamics}
\author{
    Weiqi Li \textsuperscript{\rm 1}, Zehao Zhang \textsuperscript{\rm 2}, Liang Lin \textsuperscript{\rm 1} \textsuperscript{\rm 3} \textsuperscript{\rm 4}, Guangrun Wang \textsuperscript{\rm 1} \textsuperscript{\rm 3} \textsuperscript{\rm 4} \footnotemark[2]
    \\
    \textcolor{red}{https://liwq229.github.io/humangenesis}
}
\affiliations{
    \textsuperscript{\rm 1}Sun Yat-sen University,
    \textsuperscript{\rm 2} Yonsei University,
    \textsuperscript{\rm 3} X-Era AI Lab,
    \textsuperscript{\rm 4} Guangdong Key Laboratory of Big Data Analysis and Processing
}

\begin{document}

\twocolumn[{%
\renewcommand\twocolumn[1][]{#1}%
\maketitle
\begin{center}
    \centering
    \captionsetup{type=figure}
    \includegraphics[width=0.9\textwidth]{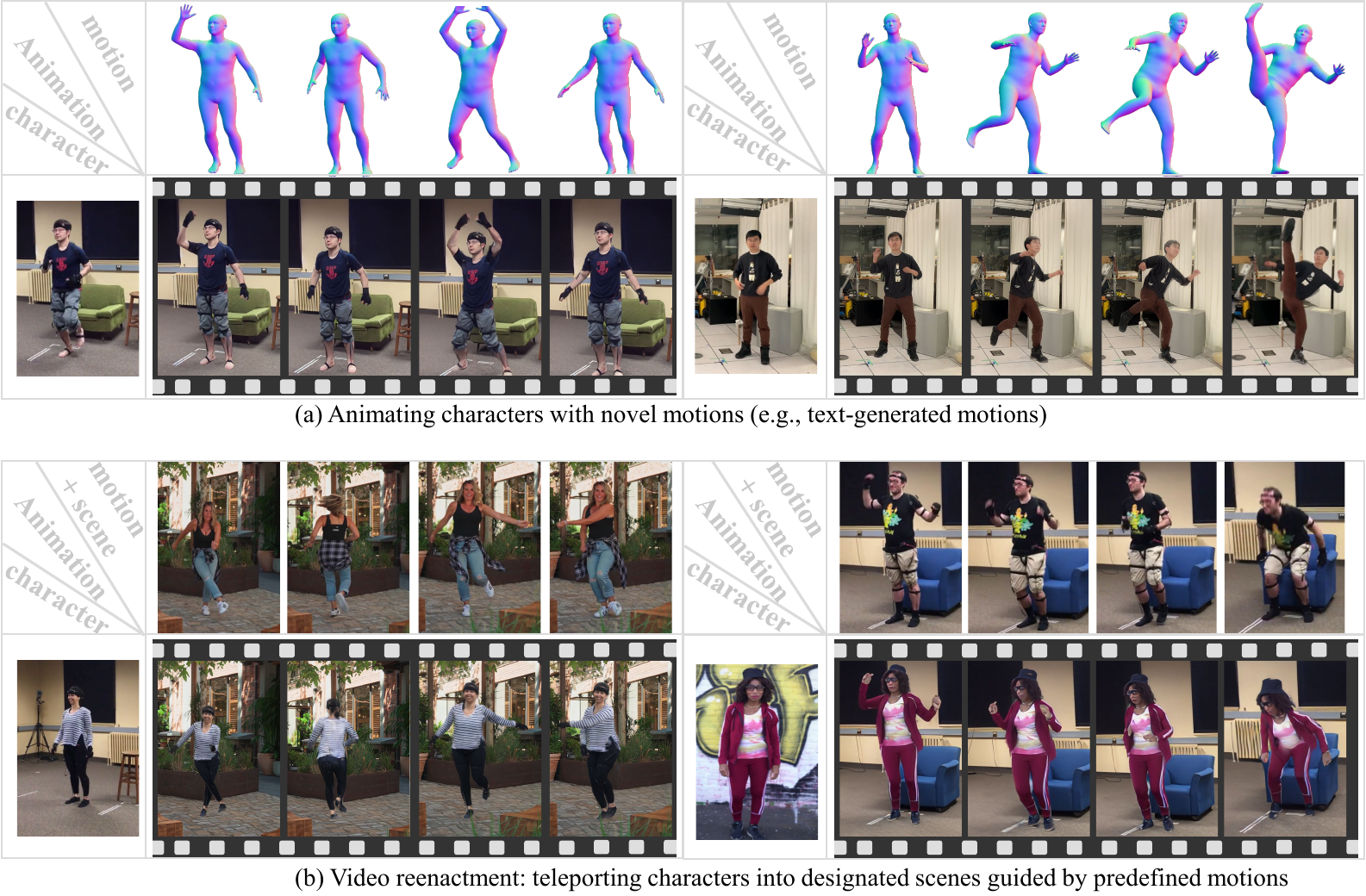}
    \captionof{figure}{
     \textbf{Compelling applications} of our proposed method in human dynamics synthesis highlight four key strengths: temporal consistency, geometric plausibility, expressive motion handling, and seamless integration into target scenes. These capabilities enable a wide range of applications, including but not limited to: (a) animating characters with novel motions—either drawn from motion capture datasets such as AMASS~\cite{AMASS:ICCV:2019} or synthesized from textual descriptions~\cite{guo2024momask}; and (b) video reenactment, where characters are smoothly inserted into target scenes and animated according to predefined motion trajectories.
    } 
    \label{fig:f1}
\end{center}
}]

\renewcommand{\thefootnote}{\fnsymbol{footnote}} 
\footnotetext[2]{Corresponding author.} 

\begin{abstract}
\textbf{Synthetic human dynamics} aims to generate photorealistic videos of human subjects performing expressive, intention-driven motions. However, current approaches face two core challenges: (1) \emph{geometric inconsistency} and \emph{coarse reconstruction}, due to limited 3D modeling and detail preservation; and (2) \emph{motion generalization limitations} and \emph{scene inharmonization}, stemming from weak generative capabilities. To address these, we present \textbf{HumanGenesis}, a framework that integrates geometric and generative modeling through four collaborative agents:
(1) \textbf{Reconstructor} builds 3D-consistent human-scene representations from monocular video using 3D Gaussian Splatting and deformation decomposition.
(2) \textbf{Critique Agent} enhances reconstruction fidelity by identifying and refining poor regions via multi-round MLLM-based reflection.
(3) \textbf{Pose Guider} enables motion generalization by generating expressive pose sequences using time-aware parametric encoders.
(4) \textbf{Video Harmonizer} synthesizes photorealistic, coherent video via a hybrid rendering pipeline with diffusion, refining the Reconstructor through a Back-to-4D feedback loop.
HumanGenesis achieves state-of-the-art performance on tasks including text-guided synthesis, video reenactment, and novel-pose generalization, significantly improving expressiveness, geometric fidelity, and scene integration.
\end{abstract}
\section{Introduction}
\label{sec:intro}

\textbf{Synthetic human dynamics} focuses on generating photorealistic videos of human subjects performing \textcolor{purple}{expressive}\footnote{Here, ``expressive'' refers to motions that convey realistic intentions and emotions, appearing lifelike, believable, and physically natural.} and intention-driven motions. Such capabilities enable a wide range of applications, including: (1) personalized avatar animation for telepresence, (2) motion retargeting in VR/AR environments, and (3) synthetic behavioral data generation for training embodied AI agents.

Recent research predominantly relies on \emph{image-conditioned, pose-guided generation} frameworks~\cite{guo2023animatediff,Wang_Yuan_Zhang_Chen_Wang_Zhang_Shen_Zhao_Zhou_2023,ma2024followposeposeguidedtexttovideo,zhang2024mimicmotionhighqualityhumanmotion,hu2024animateanyoneconsistentcontrollable,Zeng_Wei_Zheng_Zou_Wei_Zhang_Li_2023}. These approaches project reference images and 2D pose sequences into latent spaces, which are then processed by temporal diffusion models. While such models have achieved impressive results~\cite{men2025mimocontrollablecharactervideo, hu2024animateanyoneconsistentcontrollable,zhu2024champ,shao2024human4dit}, they often fall short in modeling \emph{expressive} human dynamics, due to their limited understanding of underlying 3D geometry.
Specifically, we identify two core geometric limitations:
(1) \textbf{Geometric Inconsistency:} Pose-guided synthesis often neglects body shape and structural coherence, causing inconsistencies across frames. Moreover, reliance on 2D pose estimators~\cite{yang2023effectivewholebodyposeestimation,güler2018denseposedensehumanpose} and static images~\cite{Rombach_Blattmann_Lorenz_Esser_Ommer_2022} leads to poor 3D fidelity.
(2) \textbf{Coarse Reconstruction:} These methods fail to recover fine-grained details and spatial relationships, resulting in blurry outputs and weak human-scene integration.

On the other hand, 3D reconstruction-based approaches~\cite{Li_Zheng_Wang_Liu_2023,Qian_Wang_Mihajlovic_Geiger_Tang_2023,hu2023gauhuman,lei2023gartgaussianarticulatedtemplate,paudel2024ihumaninstantanimatabledigital,kocabas2024hugs} offer stronger geometric priors. Some even incorporate \textcolor{purple}{partial}\footnote{Here, ``partial'' means that they only focus on geometric modeling on the foreground while neglecting the background.} scene modeling~\cite{peng2021neural,chen2022relighting4dneuralrelightablehuman,qiu_anigs_2024}. However, they suffer from two critical generative limitations:
(3) \textbf{Poor Motion Generalization:} These methods struggle to generate expressive or diverse motion sequences across new views and scenes.
(4) \textbf{Scene Inharmonization:} Reconstructed humans often appear detached from their surroundings, lacking realism in scene composition and rendering.

\vspace{1ex}
To address these four fundamental challenges in both geometric and generative aspects, we propose \textbf{HumanGenesis}, a modular framework with four collaborative agents:

(1) \textit{Reconstructor} solves geometric inconsistency by recovering 3D-consistent human-scene structures from monocular videos. Leveraging 3D Gaussian Splatting, it decomposes rigid and non-rigid deformations to build animatable avatars with spatial awareness.

(2) \textit{Critique Agent} addresses coarse-grained reconstruction by locating and refining visually inconsistent regions. It operates as a multimodal LLM agent that iteratively critiques the rendered output and guides local improvements.

(3) \textit{Pose Guider} enhances motion generalization by generating time-aware parametric pose sequences across novel scenes. It encodes these into temporal embeddings to drive expressive dynamics.

(4) \textit{Video Harmonizer} eliminates scene inharmonization through a hybrid rendering pipeline that incorporates video diffusion. By forming a Back-to-4D iterative feedback loop, it also refines upstream geometric reconstruction.

Together, these agents form a tightly integrated system that bridges the gap between geometric fidelity and generative expressiveness. HumanGenesis achieves state-of-the-art results in video reenactment, novel-pose animation, and text-guided human dynamics, significantly advancing geometric plausibility, motion diversity, and scene realism.

\begin{figure*}[t]
    \centering
    \includegraphics[width=\linewidth]{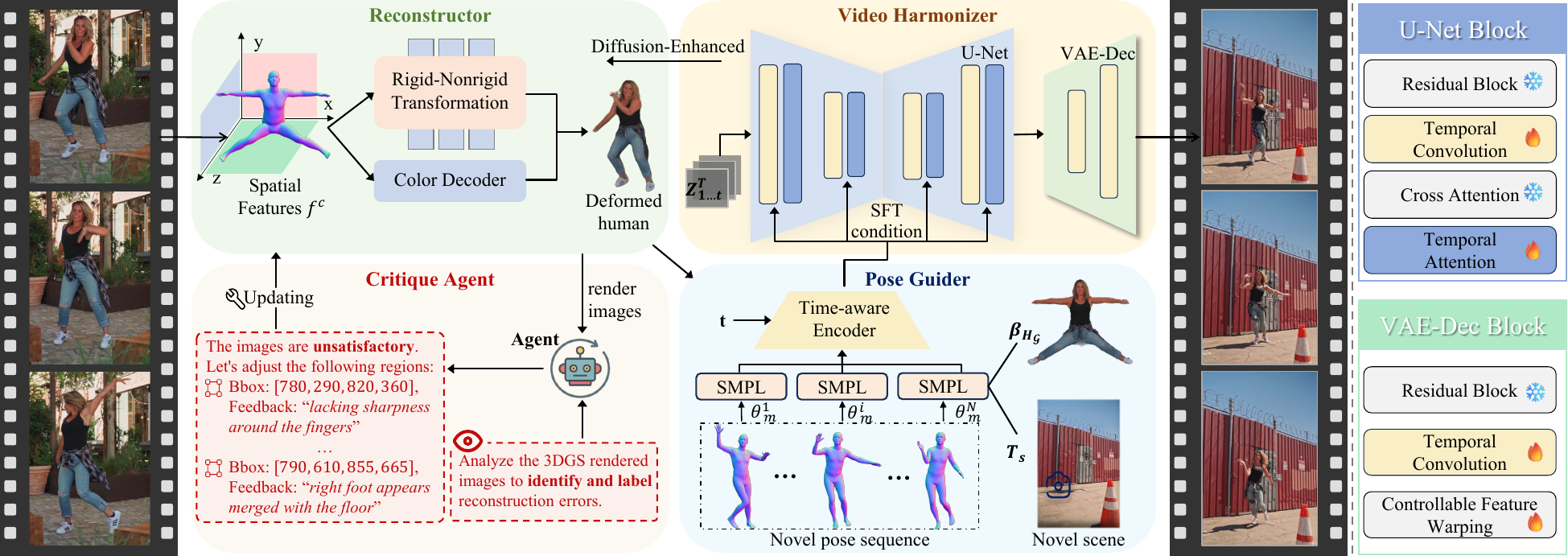} 
    \caption{{\bf Overview of HumanGenesis.} \small{The \textbf{Reconstructor} first recovers the 3D human and scene from monocular video by modeling rigid and non-rigid deformations. The \textbf{Critique Agent} then evaluates the rendered outputs to identify and refine low-quality regions, enabling fine-grained reconstruction. Next, the \textbf{Pose Guider} generates temporally-aware embeddings from novel parametric pose sequences using a time-aware encoder, allowing expressive motion synthesis. Finally, the \textbf{Video Harmonizer} leverages Spatial Feature Transform (SFT)~\cite{wang2018recoveringrealistictextureimage} within a video diffusion pipeline to produce photorealistic sequences and forms a feedback loop that enhances the Reconstructor’s input.}}
  \label{fig:framework}
\end{figure*}

\section{Related Work}
\label{sec:related_work}

\paragraph{Human-Scene Reconstruction.}
Reconstructing humans and scenes from monocular video has seen substantial progress through advances in neural implicit representations and radiance fields. Methods such as~\cite{Li_Zheng_Wang_Liu_2023,Qian_Wang_Mihajlovic_Geiger_Tang_2023,hu2023gauhuman,lei2023gartgaussianarticulatedtemplate,paudel2024ihumaninstantanimatabledigital,kocabas2024hugs} use NeRF-based models~\cite{Mildenhall_Srinivasan_Tancik_Barron_Ramamoorthi_Ng_2020} or their efficient extensions via 3D Gaussian Splatting (3DGS)~\cite{kerbl3Dgaussians,qiu_anigs_2024} to represent deformable human bodies in 3D. These models focus on geometric fidelity but often lack mechanisms for fine-grained correction or generative refinement, leading to blurred boundaries, coarse temporal consistency, or poorly integrated lighting. Moreover, most reconstructions treat the human and scene separately, limiting their spatial and visual coherence. In contrast, our \textbf{Reconstructor} and \textbf{Critique Agent} jointly address these challenges by decomposing rigid and non-rigid deformations and introducing an iterative quality-refinement loop using agential feedback. This enables fine-grained, geometry-consistent reconstructions that are further enhanced via generative correction.

\paragraph{Human Video Generation.}
Pose-guided human video generation aims to animate a target identity according to user-specified motion inputs. Recent methods~\cite{Wang_Yuan_Zhang_Chen_Wang_Zhang_Shen_Zhao_Zhou_2023,ma2024followposeposeguidedtexttovideo,zhang2024mimicmotionhighqualityhumanmotion,hu2024animateanyoneconsistentcontrollable,Zeng_Wei_Zheng_Zou_Wei_Zhang_Li_2023,Wang_Li_Lin_Zhai_Lin_Yang_Zhang_Liu_Wang_2023} generate high-quality sequences conditioned on 2D keypoints or pose sequences. For example, AnimateAnyone~\cite{hu2024animateanyoneconsistentcontrollable} introduces ReferenceNet to preserve identity and applies a pose guider for motion control. Champ~\cite{zhu2024champ}, Human4DiT~\cite{shao2024human4dit}, and MIMO~\cite{men2025mimocontrollablecharactervideo} incorporate 3D shape priors~\cite{Pavlakos_Choutas_Ghorbani_Bolkart_Osman_Tzionas_Black_2019,Loper_Mahmood_Romero_Pons-Moll_Black} to improve structural alignment. However, these methods often suffer from limited generalization to novel poses or views, especially under large-angle motion~\cite{huang2024vbench}, and can result in body deformation or identity drift.
To address these shortcomings, \textbf{HumanGenesis} introduces a \textbf{Pose Guider} that generates time-aware pose embeddings for expressive motion synthesis, and a \textbf{Video Harmonizer} that employs a pretrained video diffusion model to ensure seamless human-scene integration. This generative modeling not only enhances realism but also establishes a feedback loop that improves upstream reconstruction, enabling HumanGenesis to maintain temporal stability, geometric plausibility, and identity consistency across complex motions.

\section{Method}
\label{sec:method}

We present the full pipeline of \textbf{HumanGenesis} in Figure~\ref{fig:framework}.
Our approach leverages 3D priors for pose-guided human video synthesis, which in return facilitates more accurate 4D reconstruction through the integration of video diffusion-based priors. 
In this section, we first introduce the details of Reconstructor and Pose Guider in Section~\ref{sec:human_scene_reconstruction}. Next, we present the Critique Agent for spatial fine-grained reconstruction in Section~\ref{sec:recon_self_reflect}. Finally, the design details of the Video Harmonizer and Iterative Procedure for Enhancing 3DGS are provided in Section~\ref{sec:video_enhancement}.

Our method integrates a set of collaborative agents that combine 3D geometric priors with video diffusion-based generative refinement to synthesize expressive, pose-guided human videos in complex scenes. This agent-driven architecture enables accurate 4D human-scene reconstruction while preserving temporal consistency and photorealism.

\subsection{Human Scene Reconstruction}
\label{sec:human_scene_reconstruction}
{\bf Reconstructor.} Our Reconstructor is responsible for separating and modeling the human and static scene from a monocular video with camera motion, using 3D Gaussians. We first mask the human from the video and initialize the scene Gaussians $S_{G}$ using the structure-from-motion point cloud obtained from COLMAP~\cite{Schonberger_Frahm_2016,Schönberger_Zheng_Frahm_Pollefeys_2016}.

\noindent {\bf Non-rigid and Rigid Deformation.} Reconstructor leverages the parametric human body model, SMPL~\cite{Loper_Mahmood_Romero_Pons-Moll_Black}, 
to reconstruct a canonical 3D human avatar from the monocular video input. To model detailed elements such as hair and clothing, we employs 3D Gaussians \( H_{\mathcal{G}} \), derived from the SMPL human mesh in a Da pose, to learn human geometry and appearance in canonical space. See Appendix for details. 

To animate the human mesh into a specific pose, SMPL uses predefined joints \( n_k \) and Linear Blending Skinning (LBS), with an LBS weight matrix \( \bm{W} \in \mathbb{R}^{n_k \times n_v} \), where \( n_v \) is the number of mesh vertices \( \mathbf{x} \). Each column \( \bm{W}_i \in \mathbb{R}^{n_k} \) of \( \bm{W} \) represents the skinning weights for vertex \( \mathbf{x}_{i} \). Similarly, in our approach, given any SMPL pose parameter \( \bm{\theta} \), we can derive the bone transformation matrix \( \mathbf{B}_i \) for each joint \( i \in \{1, \ldots, n_k\} \). A 3D Gaussian point \( \mathbf{x}_d \) is then rigidly transformed to \( \mathbf{x}_p \) as follows:
\begin{equation}\label{eq:important}
    \mathbf{x}_p = \sum_{i=1}^{n_k} w^i \mathbf{B}_i \mathbf{x}_d,
\end{equation}
where \( w^i \) represents the skinning weight for joint \( i \). To learn the LBS weights for human Gaussian deformation, we draw inspiration from~\cite{hu2023gauhuman} and initialize the LBS weights $\bm{W}_c$ using SMPL. Furthermore, we utilize an LBS decoder \( \mathcal{D}_{r} \) to predict the weight offsets of the LBS, which helps to reduce computational overhead.

\noindent\textbf{Pose Guider.} Given reference video frames \( I_1, I_2, \dots, I_N \), and the pose parameters \( \bm{\theta}_m^1, \bm{\theta}_m^2, \dots, \bm{\theta}_m^N \) extracted by a pre-trained SMPL 
estimator~\cite{Goel_Pavlakos_Rajasegaran_Kanazawa_Malik_2023}, we can effectively and precisely aligns both the shape and pose of human Gaussian avatars \( H_{\mathcal{G}} \) with the motion sequence. The deformed human Gaussians ${H^{'}_{\mathcal{G}}}$ are mapped with a corresponding SMPL model as follows:
\begin{equation}
    \mathrm{SMPL}({{H^{'}_{\mathcal{G}}}}^{i}) = (\beta_{H_{\mathcal{G}}}, \bm{\theta}_m^i),
    \label{eq:important}
\end{equation}
where $\beta_{H_{\mathcal{G}}}$ is the shape parameter and \( N \) represents the total number of frames. Champ~\cite{zhu2024champ} incorporates various 2D representations of the aligned SMPL mesh to control video synthesis. 
In contrast, our method directly captures the visual appearance of human in pixel space and employ a compact, time-aware encoder to condition out Video Harmonizer.

\begin{figure}
\centering
\includegraphics[width=\linewidth]{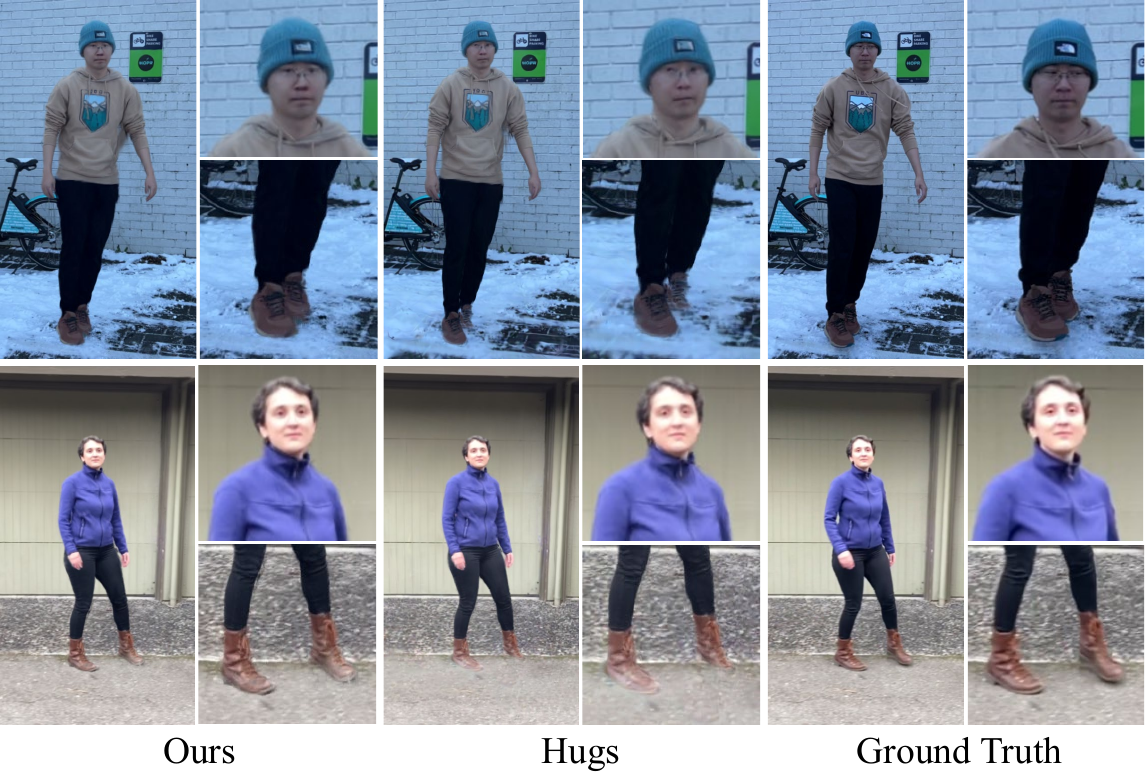}
\caption{\textbf{Visualization} of our reconstruction method. Our method effectively eliminates artifacts such as unclear separation between the feet and the floor.}
\label{fig:3d-reconstuct}
\end{figure}

\noindent \textbf{Scene Fusion.} By reconstructing the human and scene separately, our approach enables the integration of various scene settings in human video generation, which is rarely seen in previous video generation methods. Leveraging the explicit representation of 3DGS, we can incorporate new backgrounds in video generation simply by overlaying the human Gaussian ${{H_{\mathcal{G}}}}$ onto various scene Gaussians $S_{\mathcal{G}}$ with a scene translation $T_s$ and performing volumetric rendering on the combined structure.

\subsection{Critique Agent}
\label{sec:recon_self_reflect}
\textbf{Spatial Fine-Grained Reconstruction.} Previous methods often rely on global image supervision during training, which can result in artifacts in localized areas of the human body. These artifacts may manifest as blurry or distorted limbs and poor separation between the feet and the ground plane. A crucial question is how to direct more focused attention to these specific local regions during the optimization process. To address this, we introduce a novel, multi-round self-reflection phase designed to automatically identify and refine poorly reconstructed regions.

\begin{figure}
    \centering
    \includegraphics[width=\linewidth]{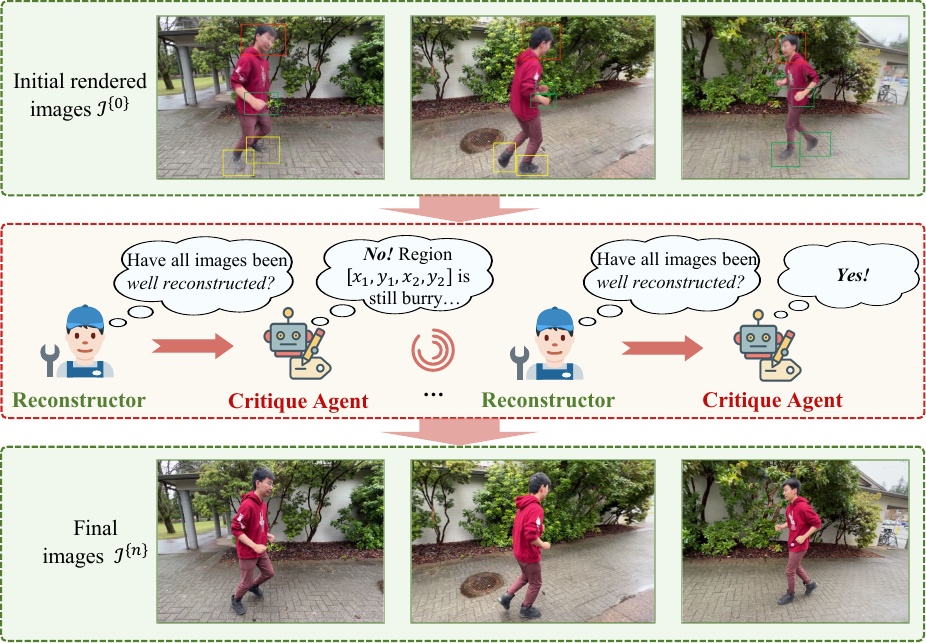} 
    \caption{\textbf{ Overview of Spatial Fine-grained Reconstruction.} We introduce a Generic AI Agent system for 3D humans reconstruction via MLLM models. The MLLM agents locate poor reconstructed regions. If any regions are flagged with negative feedback, we initiate a subsequent round of training focused specifically on these localized areas.}
  \label{fig:spatial_fine_grained_reconstruciton} 
\end{figure}

\noindent\textbf{Reconstruction with Self-Reflection.} Our approach achieves fine-grained reconstruction by emulating an expert's assessment process. The procedure commences in the initial round ($i=1$) by rendering a set of images, $\mathcal{I}^{(i)}$, from the training viewpoints based on the initial reconstruction in Section~\ref{sec:human_scene_reconstruction}. Leveraging prior knowledge of common 3DGS reconstruction failures—such as blurriness, unnatural body-part intersections, and ground fusion artifacts—the MLLM (e.g., Qwen-VL~\cite{wang2024qwen2vlenhancingvisionlanguagemodels}) localizes these poorly reconstructed regions and outputs a corresponding feedback label for each,
\begin{equation}
    (b_{j}^{(i)}, f_{j}^{(i)})  = \mathrm{MLLM}(\mathcal{I}^{(i)})
\end{equation}
\noindent where $b_{j}^{(i)}$ represents the set of identified regions, specified by 2D bounding boxes, and $f_{j}^{(i)}$ is the corresponding set of textual feedback labels for those regions. 

If all feedback labels in the set indicate that the regions are ``\textit{well reconstructed}", the process concludes and the reconstruction from this round is deemed final. In contrast, if any regions are flagged with negative feedback, we initiate a subsequent round of training focused specifically on these localized areas. This process continues until a satisfactory result is achieved. To avoid an infinite loop, we establish a maximum number of rounds. If this threshold is exceeded, the cycle terminates and we select the reconstruction from the round that yielded the smallest set of poorly reconstructed regions. We also introduce a filtering mechanism to discard inputs with SMPL estimate errors for robust reconstruction. See detailed MLLM prompts in Appendix.

\begin{figure*}[!t]
  \centering
  \includegraphics[width=0.85\linewidth]{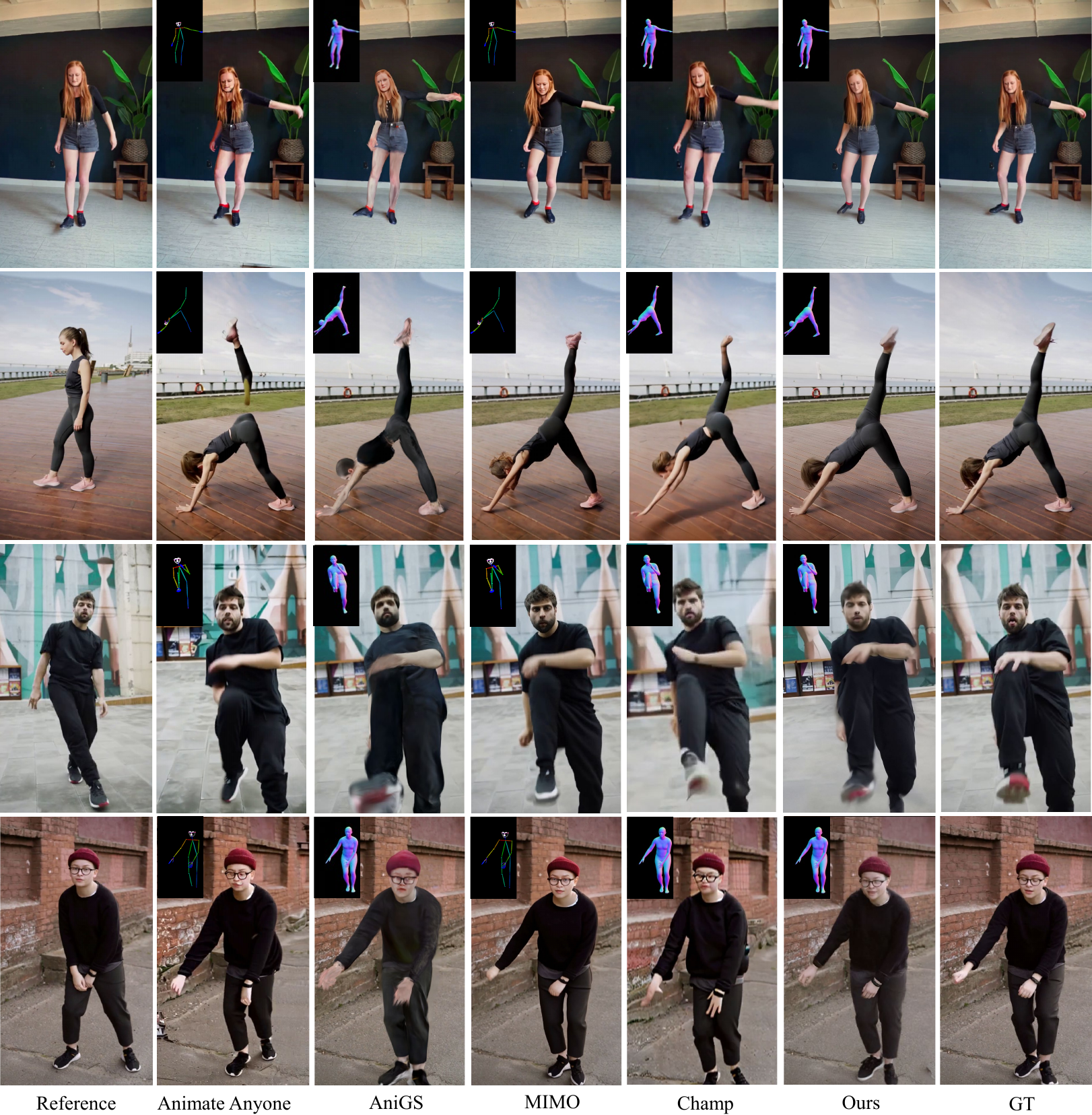} 
  \caption{\textbf{Qualitative comparisons} with state-of-the-art methods on the HumanVid dataset demonstrate that our approach consistently delivers high-quality results, even in the presence of large motions. Please zoom in to better observe the detailed differences.}
  \label{fig:main}
  \vspace{-2ex}
\end{figure*}

\subsection{Video Harmonizer}
\label{sec:video_enhancement}
\textbf{HumanGenesis} aims to generate high-quality pose-guided human videos set in diverse 3D scenes from monocular videos. 
Leveraging the inherent 3D nature of volumetric rendering, our Reconstructor ensures geometric integrity throughout the video generation process. However, render-guided video generation introduces challenges, such as the loss of detailed body features or distortion of scenes, the disharmony between the human and the scene (e.g., in terms of illumination) and temporal inconsistencies caused by Pose Guider. To address these issues, we introduce a video diffusion model trained on human video sequences.

\noindent\textbf{Spatial-Temporal Aware Video Harmonizer.} As illustrated in Figure~\ref{fig:framework}, our approach extends the existing U-Net and VAE-Decoder architecture to handle the unique challenges of human video synthesis in complex 3D environments. The temporal U-Net incorporates both spatial and temporal dimensions to effectively capture and model the dynamic motion of humans across video frames, ensuring that the generated videos maintain temporal consistency and smooth transitions. By introducing temporal attention mechanisms~\cite{Blattmann_Rombach_Ling_Dockhorn_Kim_Fidler_Kreis_2023, wang2023videocomposercompositionalvideosynthesis, ge2024preservecorrelationnoiseprior,yang2023rerendervideozeroshottextguided}  and 3D convolution blocks ~\cite{Blattmann_Rombach_Ling_Dockhorn_Kim_Fidler_Kreis_2023, wang2023videocomposercompositionalvideosynthesis} into the U-Net, we enhance the model's ability to process motion sequences with high fidelity, while maintaining the geometric accuracy of the human body in 3D space. 

The 3D-rendered conditions encoded by Pose Guider are then injected into the denoising U-Net through a spatial feature transform operation~\cite{wang2018recoveringrealistictextureimage}. To preserve temporal dynamics in the rendered sequence, a Motion-guided Diffusion Sampling strategy~\cite{yang_motion-guided_2024} is employed. This strategy leverages optical flow to align features across frames, ensuring temporal consistency through motion-guided loss.

\begin{table}[!ht]
  \centering
      \resizebox{.48\textwidth}{!}{
        \begin{tabular}{c|ccc|cc}
        \toprule
        Method & PSNR $\uparrow$ & SSIM $\uparrow$ & LPIPS $\downarrow$ & FID-VID $\downarrow$ & User $\uparrow$ \\
        \midrule
        AnimateAnyone & 16.102 & 0.633 & 0.398 & 99.49 & 2.482 \\
        AniGS & 18.974 & 0.649 & 0.373 & 86.62 & 3.136 \\
        MIMO  & 20.845 & 0.783 & 0.266 & 82.34 & 3.504 \\
        Champ & 17.986 & 0.651 & 0.382 & 101.91 & 2.637 \\
        Ours  & \textbf{21.934} & \textbf{0.802} & \textbf{0.223} & \textbf{76.20} & \textbf{4.103} \\
        \bottomrule
        \end{tabular}
        
        }
        \centering \caption{\textbf{Quantitative comparisons} with video synthesis models on HumanVid~\cite{wang2024humanviddemystifyingtrainingdata} dataset.}
      \label{tab:main}
        
\end{table}

\begin{table}[ht]
      \centering
        \begin{tabular}{c|ccc}
        \toprule
        Method & SSIM $\uparrow$  & PSNR $\uparrow$  & LPIPS $\downarrow$ \\
        \midrule
        Vid2Avatar & 0.566 & 16.192 & 0.637 \\
        NeuMan & 0.793 & 24.543 & 0.265 \\
        HUGS  & 0.829 & 25.498 & 0.122 \\
        Ours  & \textbf{0.891} & \textbf{26.992} & \textbf{0.081} \\
        \bottomrule
        \end{tabular}
        \caption{\textbf{Quantitative comparisons} with 3D reconstruction methods on NeuMan~\cite{Jiang_Yi_Samei_Tuzel_Ranjan} dataset.}
      \label{tab:3d}
        \vspace{-2ex}
\end{table}

\noindent\textbf{Temporal VAE-Decoder Fine-tuning.} Operating in a compressed latent space (8× smaller than the original image space) often leads to temporal inconsistencies in video details using conventional VAE decoders. To address this, we introduce a temporal-aware decoding framework that enhances interframe consistency through a sequence-aware module. As shown in Figure~\ref{fig:framework}, we integrate temporal 3D residual blocks and finetune using a hybrid loss, including $\mathcal{L}_{1}$ loss, LPIPS perceptual loss~\cite{johnson2016perceptuallossesrealtimestyle} and GAN loss from a patchwise temporal discriminator~\cite{zhou2023propainterimprovingpropagationtransformer}. We further incorporate the frame difference loss \(L_{\mathrm{diff}}\)~\cite{yang_motion-guided_2024} into the total loss:
\begin{equation}
    \mathcal{L}_{\mathrm {total}} = \mathcal{L}_{1} + \mathcal{L}_{LPIPS} + \alpha \mathcal{L}_{\mathrm{diff}} + \beta \mathcal{L}_{\mathrm{GAN}}
\end{equation}
where $\alpha=0.5$ and $\beta=0.025$.
Additionally, we incorporate information from the VAE encoder using a Controllable Feature Warping (CFW) module~\cite{zhou2022robustblindfacerestoration,wang2024exploitingdiffusionpriorrealworld}, which improves both restoration and generation quality. To preserve the integrity of the pre-trained VAE, we freeze the spatial layers of the encoder and only update the parameters in the temporal convolution and CFW modules.

\noindent\textbf{Back-to-4D: 4D Reconstruction with Video Diffusion Priors.} We introduce an Iterative Procedure for Enhancing 3DGS with Video Diffusion Priors. Constructing an articulated human model from a video often suffers from incomplete coverage of the input, leading to potential gaps in reconstruction. Inspired by \cite{haque2023instructnerf2nerfediting3dscenes}, we propose an iterative procedure to enhance Human Gaussian representations using priors derived from our Video Harmonizer. In each iteration, we first generate a set of synthetic views with novel camera poses or trajectories based on the Human-Scene 3DGS reconstructed in Section~\ref{sec:human_scene_reconstruction}. These synthetic views are then denoised using our trained Video Harmonizer. The resulting enhanced outputs are used to refine and update the 3DGS, incorporating additional details and improving consistency. This refinement process takes approximately 2.5k training iterations. Unlike the distillation process proposed in \cite{haque2023instructnerf2nerfediting3dscenes}, which may result in inconsistent edits, our approach ensures that the output maintains a coherent and temporally consistent representation of the human-scene 3DGS.

\section{Experiments}
\label{experiments}
\subsection{Implementations}
{\bf Implementation Details.} The denoising U-Net in our HumanGenesis is trained on 8 NVIDIA A100-80G GPUs with a batch size of 48. We initialize the denoising U-Net weights using Stable Diffusion V2.1 and extend the 2D convolutions to 3D to capture temporal dynamics. The latent image size and sequence length are set to \( 64 \times 64 \) and 6. See Appendix for more details. As for the VAE Decoder, we train it with the latent sequence sampled by finetuned U-Net and GT sequence. The training batch size and sequence length are set to 8 and 5. 
 To generate outputs at arbitrary resolutions, we utilize a progressive patch aggregation sampling strategy~\cite{wang2024exploitingdiffusionpriorrealworld}, running the sampling process for 50 steps per sequence.

\noindent {\bf Dataset.} 
Everyday videos frequently include camera motion, which facilitates our 3D human reconstruction process. We curated an in-the-wild monocular video dataset from online repositories. The MoVi dataset~\cite{Ghorbani_Mahdaviani_Thaler_Kording_Cook_Blohm_Troje} provides 225 videos with distinct individuals performing full-body motions in indoor settings. Additionally, we collected 1K single-person videos from HumanVid~\cite{wang2024humanviddemystifyingtrainingdata}, covering a range of ages and capturing various perspectives, including full-body, half-body, and close-up shots, in both indoor and outdoor environments.

\begin{figure}[t]
    \centering
    \includegraphics[width=\linewidth]{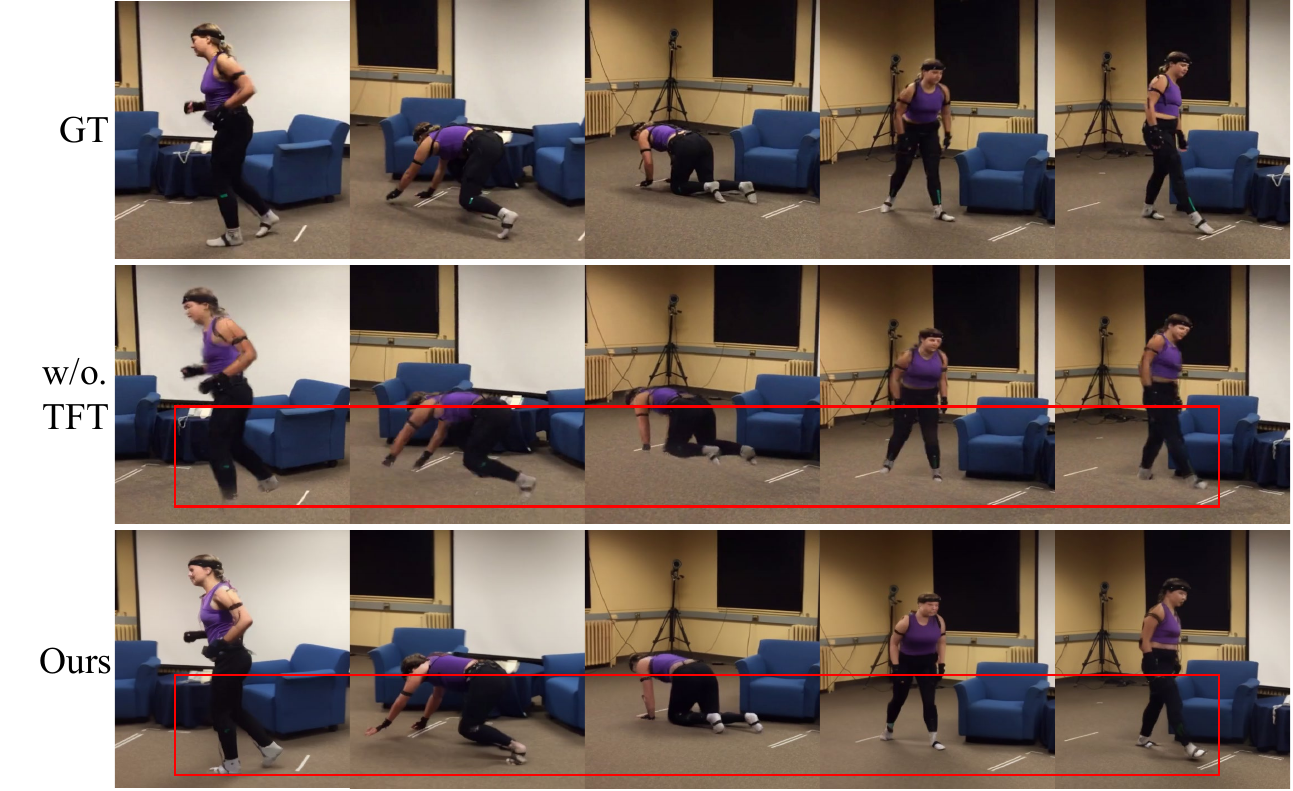}
    \captionof{figure}{\textbf{Qualitative ablation analysis.}``w/o. TFT" denotes results generated without Temporal Fine-Tuning (TFT) applied to the U-Net. Please refer to the supplementary video for further details.}
    \label{fig:TFT}
\end{figure}

\subsection{Comparisons}

In this section, we present a comparison of our method with prior state-of-the-art approaches, namely Animate Anyone\footnote{https://github.com/MooreThreads/Moore-AnimateAnyone}~\cite{hu2024animateanyoneconsistentcontrollable}, AniGS~\cite{qiu_anigs_2024}, MIMO~\cite{men2025mimocontrollablecharactervideo}, and Champ~\cite{zhu2024champ}. As Animate Anyone is not open-source, we use a third-party implementation for this method, while the official implementations are applied for
the other methods. Note that these approaches above only use a single input reference image. Therefore, We split the monocular video into multiple sequences with a stride of 10. Our Reconstructor take all first frames of subsequences as the training set, while other methods sample each subsequence with the first frame of the subsequence as the reference image. All methods are then evaluated on the test frames, and we assess each video using the PSNR~\cite{Wang_Bovik_Sheikh_Simoncelli_2004}, SSIM~\cite{Wang_Bovik_Sheikh_Simoncelli_2004}, LPIPS~\cite{Zhang_Isola_Efros_Shechtman_Wang_2018} and FID~\cite{Heusel_Ramsauer_Unterthiner_Nessler_Hochreiter_2017} metrics.

\noindent{\bf Qualitative Comparison.} We evaluate the qualitative performance of our method, HumanGenesis, and its competitors. Figure~\ref{fig:main} illustrates the results of different pose-driven video generation methods applied to complex or large-scale human motions. 
AniGS reconstructs a consistent human avatar using 4DGS but fails to account for scene illumination, resulting in unrealistic visuals. Champ incorporates multiple 2D motion signals, projected by a shape-aligned SMPL model, but still suffers from artifacts and unrealistic results, particularly in facial rendering. While MIMO also decomposes the 2D video in 3D-aware manner, multi-faceted encoding mechanisms may compromise the stability of diffusion-based generation, particularly when character performs large-angle rotations, as evidenced by the face inconsistency observed in the first and third row of Figure~\ref{fig:main}. In contrast, Reconstructor provides a more reliable 3D reference, ensuring consistent appearance. Besides, when handling large-scale motions, Pose Guider in HumanGenesis effectively aligns the pose, maintaining consistency even during complex movements. 

\begin{table}[t]
  \centering
  \small
    \begin{tabular}{c|ccc}
    \toprule
    Setting & SSIM $\uparrow$ & PSNR $\uparrow$ & LPIPS $\downarrow$\\
    \midrule
    
    Initial & 0.842  & 26.723  & 0.115  \\
    Round 1 & 0.913 & 28.254  & 0.070  \\
    Round 2 & 0.914  & 28.348  & \textbf{0.068} \\
    Round 3 & \textbf{0.915} & \textbf{28.414} & \textbf{0.068} \\
    Round 4 & 0.914  & 28.386  & \textbf{0.068} \\
    \bottomrule
    \end{tabular}
    \caption{\textbf{Ablation study} on the number of reflection rounds. We evaluate Critique Agent's performance across multiple rounds on Neuman dataset.}
  \label{tab:ablation_1}
\end{table}

\begin{table}[t]
    \centering
    \small
    \begin{tabular}{c|ccc}
    \toprule
    Method & SSIM $\uparrow$ & PSNR $\uparrow$ & LPIPS $\downarrow$\\
    \midrule
    
    Ours(w/o. CFW) & 0.848 & 25.692 & 0.107 \\
    Ours(w/o. TC) & 0.821 & 25.311 & 0.141 \\
    Ours(w/o. TFT) & 0.816 & 24.978 & 0.155 \\
    Ours(w/o. IDR) & 0.855 & 26.103 & 0.102 \\
    Ours  & \textbf{0.873} & \textbf{26.335} & \textbf{0.092} \\
    \bottomrule
    \end{tabular}
    \captionof{table}{\textbf{Ablation study} on Video Harmonizer and Iterative Diffusion-enhanced Reconstruction.}
    \label{tab:ablation}
    \vspace{-2ex}
\end{table}

\begin{figure}[t]
  \centering
  \includegraphics[width=\linewidth]{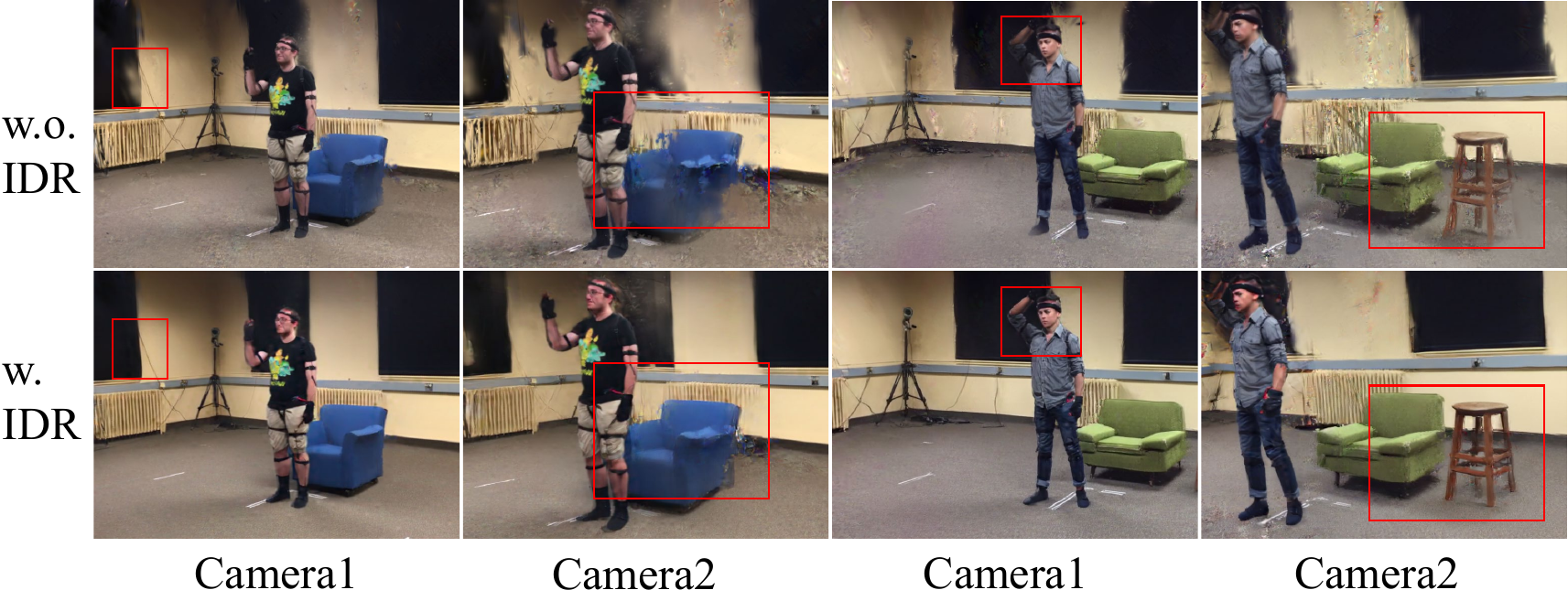}
  \captionof{figure}{\textbf{Qualitative ablation studies} demonstrate the impact of Video Harmonizer, where ``w. IDR" denotes results from novel camera poses after Iterative Diffusion Reconstruction via Temporal Fine-Tuning U-Net.}
  \label{fig:ablation}
\end{figure}
\noindent{\bf Quantitative Comparison.} As shown in Table~\ref{tab:main}, for complex and large-scale motion datasets with camera movement, the proposed method demonstrates superior performance, achieving higher PSNR and SSIM values, while significantly reducing LPIPS and FID-VID scores. We also conducted a user study to evaluate the visual quality of the generated videos. We randomly selected 15 samples and organized anonymous rankings from 25 participants, with the highest score being 5. After averaging the scores, we obtained the final metric. The user study results are consistent with the quantitative evaluation, further validating the perceptual quality of our generated videos.
Table~\ref{tab:3d} shows a comparison on NeuMan~\cite{Jiang_Yi_Samei_Tuzel_Ranjan} dataset between our spatial fine-grained reconstruction method and other approaches. Figure~\ref{fig:3d-reconstuct} shows our method outperforms these methods, achieving superior visual details.

\subsection{Ablation Studies}

To validate the effectiveness of \textbf{Critique Agent}, we conduct an ablation study on the multi‑round reflection setting. We systematically evaluate the impact of reflection rounds on reconstruction performance using the Neuman dataset. As shown in Table~\ref{tab:ablation_1}, with an increasing number of reflection rounds, Critique Agent demonstrates significant and consistent improvements across key metrics. Notably, after reaching the optimal reflection round (Round 3), the performance stabilizes, with no further meaningful gains observed. These results clearly indicate that the multi‑round reflection mechanism effectively identifies and refines reconstruction errors in fine‑grained regions.

\noindent \textbf{Video Harmonizer and Diffusion-Enhanced Reconstruction.} Consider that 3D reconstruction focuses on restoring the geometric pixel values of the scene as accurately as possible, but it does not address temporal consistency or human-scene harmonization. Hence, we incorporate a video diffusion model to generate more stable and coherent videos. Figure~\ref{fig:TFT} and Table~\ref{tab:ablation} presents the results of our ablation study on Video Harmonizer, conducted on the MoVi dataset, showing consistent improvements in all the metrics evaluated. Here, TC, TFT and IDR refer to Temporal Convolution, Temporal Fine-Tuning, and Iterative Diffusion-Enhanced Reconstruction, respectively. As shown in Figure~\ref{fig:ablation}, when rendering in novel camera poses, our proposed approach significantly fixes artifacts in human scene reconstruction. See Appendix for more details.

\section{Conclusion}
\label{sec:conclusion}

We introduced \textbf{HumanGenesis}, a collaborative agent-based framework for synthesizing photorealistic and expressive human videos. By unifying geometric reconstruction with generative refinement, our method achieves strong performance in motion expressiveness, geometric fidelity, and scene coherence. Please refer to the appendix for a discussion of limitations.

\bibliography{aaai2026}

\onecolumn
\appendix

\section{Implementation Details}  
\label{sec:supp}
\subsection{Human Gaussian Reconstructor}
\label{sec:Appendix B1}
For the fast optimization of human Gaussian, we initialize 3D Gaussians by sampling 50K SMPL vertex points. We utilize the novel adaptive control strategy in \cite{hu2023gauhuman} for 3D Gaussians. To model complex human deformation, we decouple it into a non-rigid deformation and a rigid transformation, as described in Section~\ref{sec:human_scene_reconstruction}. Specifically, our module uses a feature encoder~\cite{chan2022efficientgeometryaware3dgenerative,peng2020convolutionaloccupancynetworks} \( \bm{F} \in \mathbb{R}^{3 \times h \times w \times d} \), which encodes a feature embedding \( f^{c} \in \mathbb{R}^{d} \) for any point \( \mathbf{x}_c \) in a 3D space. A deformation decoder \( \mathcal{D}_{nr} \)  is then used to predict non-rigid deformation of 3D Gaussian, which predicts the offsets of the Gaussian’s position, scale, and rotation:
\begin{equation} 
    (\Delta x, \Delta s, \Delta \mathbf{q}) = \mathcal{D}_{nr}(f^{c})
\end{equation}
Using these predictions, we obtain the deformed canonical Gaussian:
\begin{align}
    \mathbf{x}_d&=\mathbf{x}_c+\Delta x \\
    \mathbf{s}_d&=\mathbf{s}_c\cdot\exp(\Delta s) \\
    \mathbf{q}_d&=\mathbf{q}_c\cdot[1,\Delta q_1,\Delta q_2, \Delta q_3]
\end{align}
where \( \mathbf{q} \) represents quaternions. Since we only consider the rotational component, we use only \( \Delta q_1 \), \( \Delta q_2 \), and \( \Delta q_3 \). 
For the non-rigid deformation decoder \( \mathcal{D}_{nr} \), we use an MLP with 3 hiddle layers of 128 dimensions which takes feature embedding \( f^{c} \in \mathbb{R}^{64} \) to decode non-rigid local deformation. Similarly, the color decoder consists of a 2-layer MLP that takes \( f^{c} \) as input and outputs opacity and spherical harmonics.

The rigid deformation decoder \( \mathcal{D}_{r} \) utilizes a 4-layer MLP of 128 dimensions to predict the offset of LBS weights with 24 dimensions (number of joints). ReLU activation is used after each layer. After optimizing and storing the LBS weights along with all 3D Gaussian attributes, our module facilitates direct animation of human Gaussians, eliminating the need to evaluate the triplane and decoders at runtime. The streamlined pre-processing and training pipeline requires merely 20 minutes in total, serving as an efficient foundation for generating high-fidelity 3D human priors that significantly improve video synthesis performance.

During optimization, the loss function is composed of 
\begin{equation}
    \mathcal{L}_{recon} = \mathcal{L}_{1} + \lambda_1 \mathcal{L}_{mask} + \lambda_2 \mathcal{L}_{SSIM} + \lambda_3 \mathcal{L}_{LPIPS}
\end{equation}
where we set $\lambda_1 = 0.5$, $\lambda_2 = \lambda_3 = 0.01$ to keep the same magnitude. All the networks are trained for a total of 10k iteration in 10 minutes on a single NVIDIA A100 GPU.

\subsection{Loss Function Details}

\textbf{$\mathcal{L}_{1}$ Discrepancy}. 
Let $ C \in \mathbb{R}^{H \times W \times 3} $ denote the reference image and $ \hat{C} $ the synthesized output. A pixel-wise $ \mathcal{L}_1 $-norm penalty is imposed:
\begin{equation}
\mathcal{L}_{\text{color}} = \| \hat{C} - C \|_1
\end{equation}

\noindent\textbf{Binary Mask Regularization}. 
To refine geometry learning, we incorporate foreground segmentation masks. Denoting $ \hat{M} $ as the differentiable volume density and $ M \in \{0,1\}^{H \times W} $ as the binary ground truth, the regularization term becomes:
\begin{equation}
\mathcal{L}_{\text{mask}} = \| \hat{M} - M \|_2
\end{equation}

\noindent\textbf{Structural Similarity Loss}. 
The SSIM metric quantifies spatial coherence between reconstructed and target frames:
\begin{equation}
\mathcal{L}_{\text{SSIM}} = \text{SSIM}(\hat{C}, C)
\end{equation}

\noindent\textbf{Perceptual Fidelity Loss}. 
We further adopt the LPIPS loss to penalize high-level feature mismatches:
\begin{equation}
\mathcal{L}_{\text{LPIPS}} = \text{LPIPS}(\hat{C}, C)
\end{equation}

\subsection{Detail Designs of Critique Agent}
\noindent\textbf{Training Implementation Details.}
During the training of the Critique Agent, we adopt a region-aware weighting mechanism to emphasize fine-grained reconstruction quality in spatially significant areas. For local regions identified as perceptually or semantically important—such as facial components or joints—the reconstruction loss $\mathcal{L}_{recon}$ is assigned a substantially higher weight relative to background or less critical regions. This prioritization enhances the model’s ability to capture subtle structural details. Formally, let $\mathcal{R}_{local}$ denote the set of high-importance pixels, the final loss is computed as:

\begin{equation}
\mathcal{L}_{\text{final}} = \mathcal{L}_{recon} + \omega \cdot\sum_{(i,j) \in \mathcal{R}_{local}} \left[ \mathcal{L}_{\text{color}}^{(i,j)} + \lambda_1 \mathcal{L}_{\text{mask}}^{(i,j)} + \lambda_2 \mathcal{L}_{\text{SSIM}}^{(i,j)} + \lambda_3 \mathcal{L}\_{\text{LPIPS}}^{(i,j)} \right]
\end{equation}

Here, $\omega \gg 1$ denotes a significantly amplified weighting factor applied to local regions to promote spatial accuracy. This strategy allows the model to deliver enhanced reconstruction performance in areas that are most critical to perceptual quality. Table~\ref{tab:neuman_human_scene} shows the detailed evaluation results on Neuman dataset~\cite{Jiang_Yi_Samei_Tuzel_Ranjan}.

\begin{figure}
    \centering
    \includegraphics[width=0.95\linewidth]{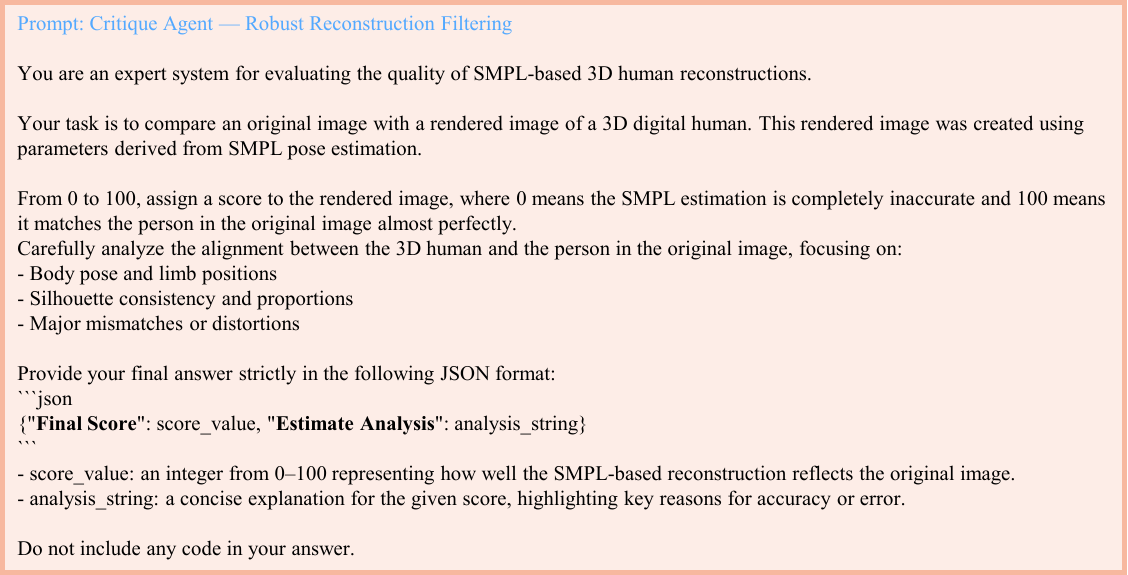}
    \caption{Prompt for Robust Reconstruction Filtering in our Critique Agent.}
    \label{fig:prompt1}
\end{figure}

\begin{figure}
    \centering
    \includegraphics[width=0.95\linewidth]{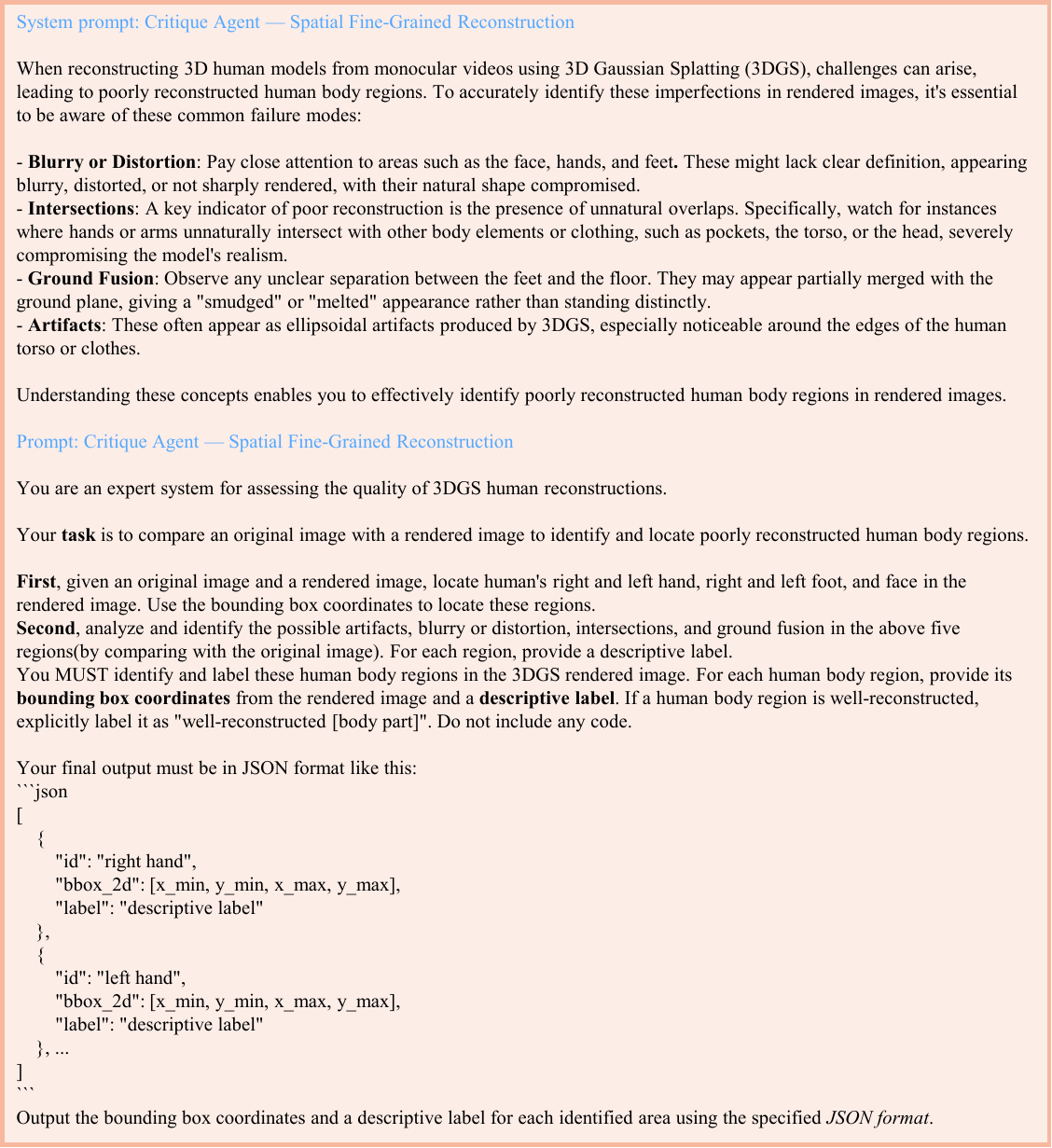}
    \caption{Prompt for Spatial Fine-Grained Reconstruction in our Critique Agent.}
    \label{fig:prompt2}
\end{figure}

\noindent\textbf{Full Prompts for Critique Agent.} Figures~\ref{fig:prompt1} and~\ref{fig:prompt2} illustrate the full textual prompts used in our Critique Agent. These prompts are specifically designed to guide the MLLM in performing two key tasks: Robust Reconstruction Filtering and Spatial Fine-grained Reconstruction. Each prompt is carefully crafted to elicit accurate and context-sensitive responses from the model, thereby enhancing the overall critique and reconstruction pipeline.

\begin{figure*}
  \centering
  \begin{subfigure}{\linewidth}
    \centering
    \includegraphics[width=\linewidth]{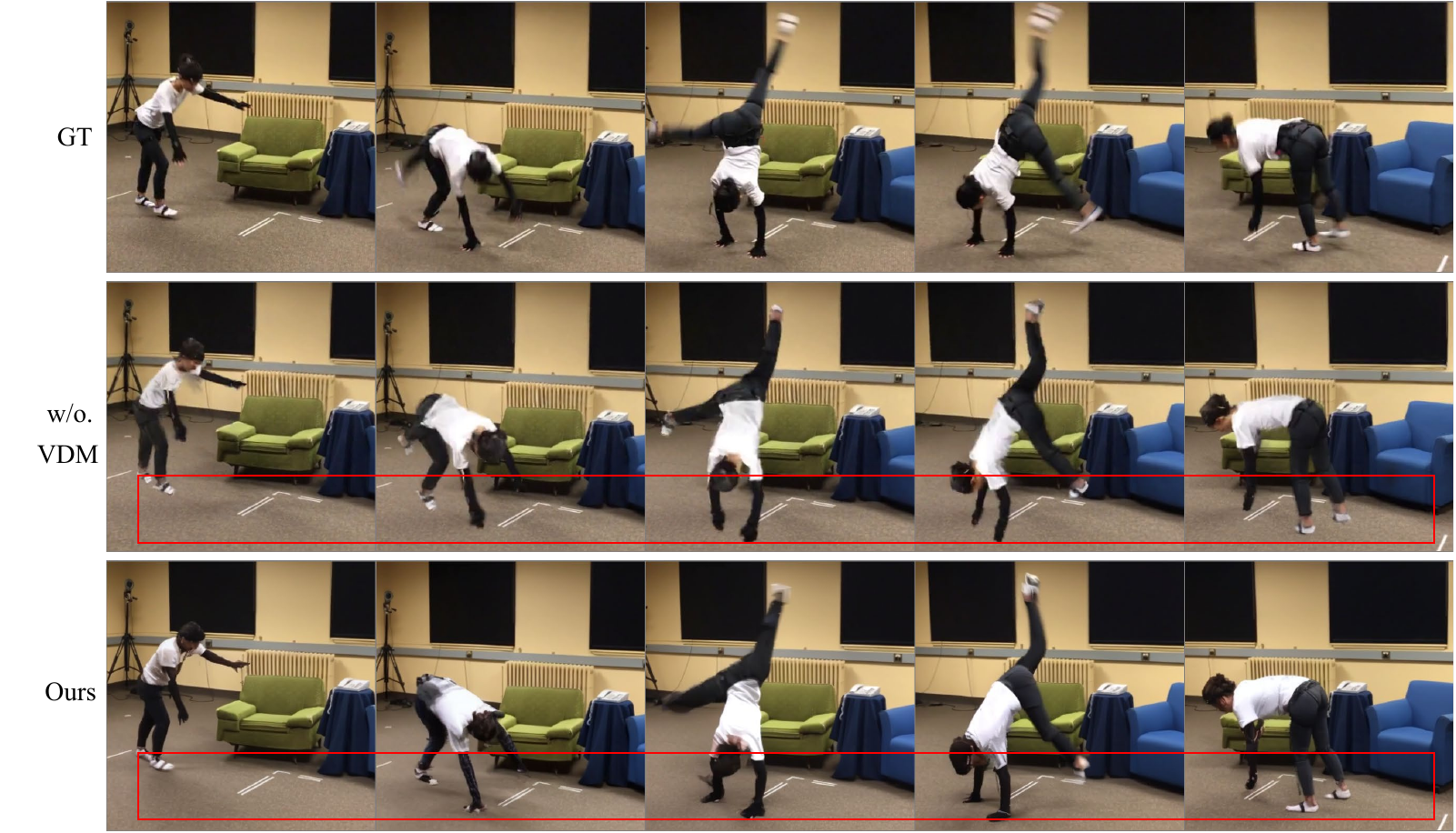} 
  \end{subfigure}
  \caption{Effectiveness of the video diffusion model under large-scale motion.}
  \label{fig:vdm2}\vspace{-2ex}
\end{figure*}

\subsection{Training Setting of Video Harmonizer}
\label{sec:Appendix B2}
We implement all models using the PyTorch\cite{paszke2019pytorchimperativestylehighperformance} framework, with Adam\cite{kingma2017adammethodstochasticoptimization} as optimizer. The temporal video diffusion model (VDM) is trained for 30,000 iterations with a learning rate of \(1 \times 10^{-3}\).  Following existing methods~\cite{Blattmann_Rombach_Ling_Dockhorn_Kim_Fidler_Kreis_2023, Zhou_Yang_Wang_Luo_Loy_2023}, the parameters of the pre-trained spatial layers are fixed and only the temporal layers are fine-tuned. Meanwhile, the VAE Decoder is first trained on the MoVi dataset for 10k iterations, followed by an additional 15k iterations on the HumanVid dataset with a learning rate of \(5 \times 10^{-5}\), ensuring improved adaptability to diverse human motions.
As described in Section~\ref{sec:video_enhancement}, we incorporate the frame difference loss \(L_{\mathrm{diff}}\)~\cite{yang_motion-guided_2024} into the total loss for decoder fine-tuning:

\begin{equation}
    \mathcal{L}_{\mathrm {total}} = \mathcal{L}_{\text{color}} + \mathcal{L}_{LPIPS} + \alpha \mathcal{L}_{\mathrm{diff}} + \beta \mathcal{L}_{\mathrm{GAN}}
\end{equation}
where $\alpha=0.5$ and $\beta=0.025$.

The frame difference loss \(L_{\mathrm{diff}}\) calculates the difference between consecutive predicted frames $\hat{I}$ and ground truth frames $I^{gt}$: \begin{equation}
    L_{\mathrm{diff}} = \textstyle\sum \Vert (\hat{I}_{i+1} - \hat{I}_{i}) - (I_{i+1}^{gt} - I_{i}^{gt}) \Vert_{1}.
\end{equation}

\definecolor{tabzero}{rgb}{1, 0.7, 0.7}
\definecolor{tabfirst}{rgb}{1, 0.85, 0.7}
\definecolor{tabsecond}{rgb}{1, 1, 0.7}

\begin{table*}[]
    \centering
    \resizebox{\textwidth}{!}{
    \begin{tabular}{c|ccc|ccc|ccc|ccc|ccc|ccc}
    \toprule
        & \multicolumn{3}{c|}{\textbf{Seattle}} & \multicolumn{3}{c|}{\textbf{Citron}} & \multicolumn{3}{c|}{\textbf{Parking}} & \multicolumn{3}{c|}{\textbf{Bike}} & \multicolumn{3}{c|}{\textbf{Jogging}} & \multicolumn{3}{c}{\textbf{Lab}}   \\
    \midrule
        & PSNR $\uparrow$ & SSIM $\uparrow$ & LPIPS $\downarrow$ & PSNR $\uparrow$ & SSIM $\uparrow$ & LPIPS $\downarrow$ & PSNR $\uparrow$ & SSIM $\uparrow$ & LPIPS $\downarrow$ & PSNR $\uparrow$ & SSIM $\uparrow$ & LPIPS $\downarrow$ & PSNR $\uparrow$ & SSIM $\uparrow$ & LPIPS $\downarrow$ & PSNR $\uparrow$ & SSIM $\uparrow$ & LPIPS $\downarrow$  \\
    \midrule    
    Vid2Avatar &  17.41 &  0.56 &  0.60 &  14.32 &  0.62 &  0.65 &  21.56 &  0.69 &  0.50 &  14.86 &  0.51 &  0.69 &  15.04 &  0.41 &  0.70 &  13.96 &  0.60 &  0.68 \\
    NeuMan     & \cellcolor{tabsecond}23.99 & \cellcolor{tabsecond}0.78 & \cellcolor{tabsecond}0.26 & \cellcolor{tabfirst}24.63 & \cellcolor{tabzero}0.81 & \cellcolor{tabsecond}0.26 & \cellcolor{tabsecond}25.43 & \cellcolor{tabsecond}0.80 & \cellcolor{tabsecond}0.31 &  \cellcolor{tabsecond}25.55 & \cellcolor{tabsecond}0.83 & \cellcolor{tabsecond}0.23 & \cellcolor{tabsecond}22.70 & \cellcolor{tabsecond}0.68 & \cellcolor{tabsecond}0.32 & \cellcolor{tabsecond}24.96 & \cellcolor{tabsecond}0.86 & \cellcolor{tabsecond}0.21 \\
    HUGS       &  \cellcolor{tabfirst}26.48    &  \cellcolor{tabfirst}0.85 &  \cellcolor{tabfirst}0.10    &  \cellcolor{tabsecond}24.35    &  \cellcolor{tabsecond}0.80    &  \cellcolor{tabfirst}0.12 &  \cellcolor{tabfirst}26.96 &  \cellcolor{tabfirst}0.84 &  \cellcolor{tabfirst}0.13 & \cellcolor{tabfirst}25.76 &  \cellcolor{tabfirst}0.84 &  \cellcolor{tabfirst}0.10 &  \cellcolor{tabfirst}23.94 &  \cellcolor{tabfirst}0.76 &  \cellcolor{tabfirst}0.20 &  \cellcolor{tabfirst}25.50 &  \cellcolor{tabfirst}0.90 &  \cellcolor{tabfirst}0.08 \\
    \midrule
    Ours       &  \cellcolor{tabzero}28.57    &  \cellcolor{tabzero}0.93 &  \cellcolor{tabzero}0.05     &  \cellcolor{tabzero}24.85    &  \cellcolor{tabzero}0.81 &  \cellcolor{tabzero}0.10 &  \cellcolor{tabzero}27.94 &  \cellcolor{tabzero}0.90 &  \cellcolor{tabzero}0.09 & \cellcolor{tabzero}27.51 &  \cellcolor{tabzero}0.91 &  \cellcolor{tabzero}0.05 &  \cellcolor{tabzero}25.86 &  \cellcolor{tabzero}0.86 &  \cellcolor{tabzero}0.13 &  \cellcolor{tabzero}27.31 &  \cellcolor{tabzero}0.93 &  \cellcolor{tabzero}0.06
    \\
    \bottomrule
    \end{tabular}  
    }
    \caption{Comparison of Critique Agent (ours) with previous work on test images of the NeuMan dataset~\cite{Jiang_Yi_Samei_Tuzel_Ranjan} using PSNR, SSIM and LPIPS metrics. HUGS achieves state-of-the-art performance.}
    \label{tab:neuman_human_scene}
\end{table*}

\subsection{Diffusion-Enhanced Reconstruction}
Algorithm\ref{alg:iterative_enhancement} outlines the proposed iterative framework for enhancing 3D Gaussian Splatting (3DGS) with video diffusion priors. Starting from an initial Human-Scene 3DGS reconstruction (Section \ref{sec:human_scene_reconstruction}), the method alternates between three key stages: (1) synthetic view generation from novel camera poses/trajectories to address incomplete input coverage; (2) diffusion-based denoising using a temporal-aware video diffusion model to refine geometric details and enforce temporal coherence; and (3) 3DGS update via optimization, where the enhanced views are used to iteratively refine the Gaussian attributes (positions, colors, opacities, and covariances). Each outer iteration includes an inner loop of T=2.5k steps to stabilize the 3DGS updates.

\begin{algorithm}[h]
\caption{Iterative Enhancement of Human-Scene 3DGS}
\renewcommand{\algorithmicrequire}{\textbf{Input:}}
\renewcommand{\algorithmicensure}{\textbf{Output:}}
\label{alg:iterative_enhancement}
\begin{algorithmic}[1]
\Require 
    \State Initial Human-Scene 3DGS $ \mathcal{G}_0 $ (from Section~\ref{sec:human_scene_reconstruction})
    \State Trained video diffusion model $ \mathcal{D} $
    \State Total outer iterations $ E $
    \State Inner iteration steps $ T = 2.5k $
\Ensure 
    \State Refined 3DGS $ \mathcal{G}_E $

\For{$ e = 1 $ to $ E $} \Comment{Outer training loop}
    \State \textbf{1. Synthetic View Generation:} 
    \State \quad Generate novel views $ \mathcal{V}_t $ from $ \mathcal{G}_{e-1} $ using predefined camera poses/trajectories.
    \State \textbf{2. Diffusion-Based Denoising:}
    \State \quad Denoise $ \mathcal{V}_t $ via $ \mathcal{D} $ to obtain enhanced views $ \hat{\mathcal{V}}_t $.
    
    \State \textbf{3. Inner Optimization Loop:} \Comment{Refine 3DGS over $ T $ steps}
    \State Initialize $ \mathcal{G}_{e,0} \leftarrow \mathcal{G}_{e-1} $
    \For{$ t = 1 $ to $ T $}
        \State Update $ \mathcal{G}_{e,t} $ using $ \hat{\mathcal{V}}_t $:
        \State $ \mathcal{G}_{e,t} \leftarrow \text{Optimize}(\mathcal{G}_{e,t-1}, \hat{\mathcal{V}}_t) $.
        \State $ \mathcal{L}_{recon} = \mathcal{L}_{1} + \lambda_1 \mathcal{L}_{mask} + \lambda_2 \mathcal{L}_{SSIM} + \lambda_3 \mathcal{L}_{LPIPS} $
       \State $ \mathcal{G}_{e,t} \leftarrow \arg\min_{\mathcal{G}} \mathcal{L}_{\text{total}} $
    \EndFor
    \State Set $ \mathcal{G}_e \leftarrow \mathcal{G}_{e,T} $ \Comment{Update global 3DGS}
\EndFor
\end{algorithmic}
\end{algorithm}

\section{More Visualization Results}

\begin{figure*}[!t]
  \centering
  \begin{subfigure}{0.85\linewidth}
    \centering
    \includegraphics[width=\linewidth]{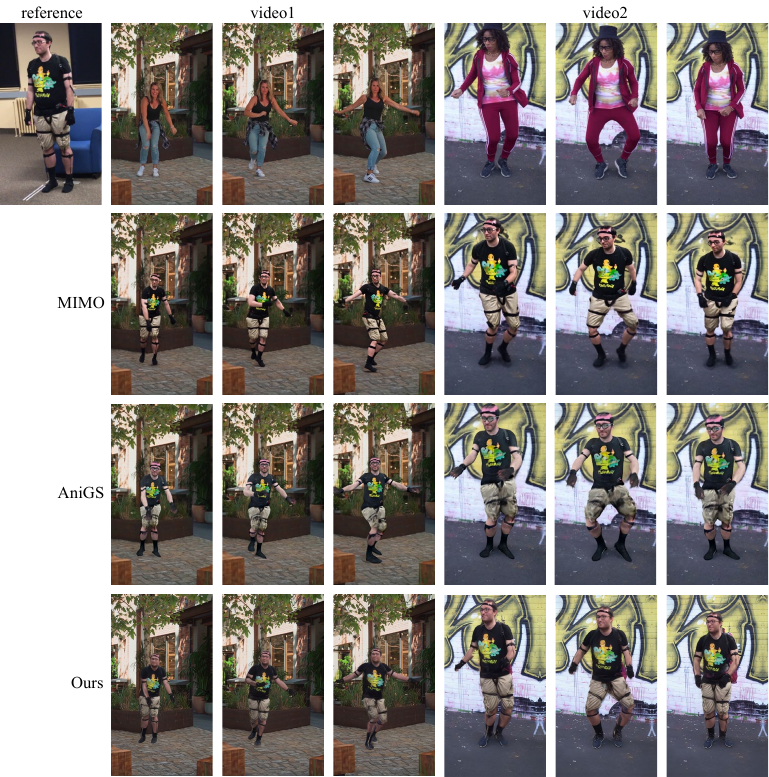} 
  \end{subfigure}
  \caption{Results of generating multiple videos of the same individual with different pose guidance.}
  \label{fig:cross_video}
\end{figure*}

\begin{figure*}[!t]
  \centering
  \begin{subfigure}{\linewidth}
    \centering
    \includegraphics[width=0.8\linewidth]{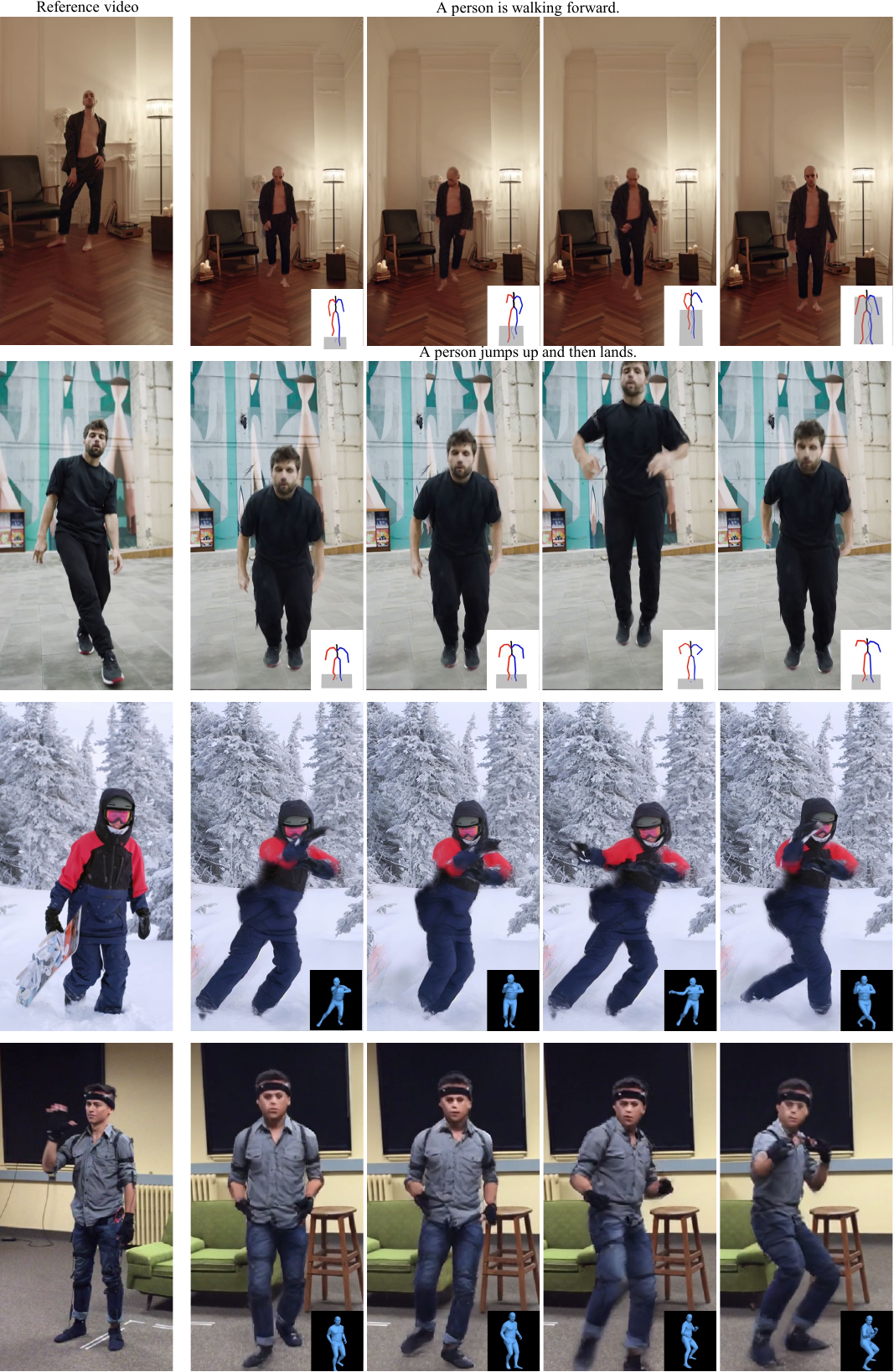} 
  \end{subfigure}
  \caption{Novel 3d pose generated by a text-to-motion approach MoMask or obtained from AMASS dataset.}
  \label{fig:suppl1}
\end{figure*}

\noindent\textbf{Video Diffusion Model.} Figure~\ref{fig:ablation} demonstrates the implicit modeling capabilities of the video diffusion model, particularly in terms of illumination and scene harmonization. We present additional visualizations of the video model under large-scale motion in Figure~\ref{fig:vdm2}. Even in scenarios with large-scale motion, our model is capable of preserving human body details under realistic lighting conditions and ensuring natural alignment between the human and the scene. 

\begin{figure*}[!t]
  \centering
  \begin{subfigure}{0.8\linewidth}
    \centering
    \includegraphics[width=\linewidth]{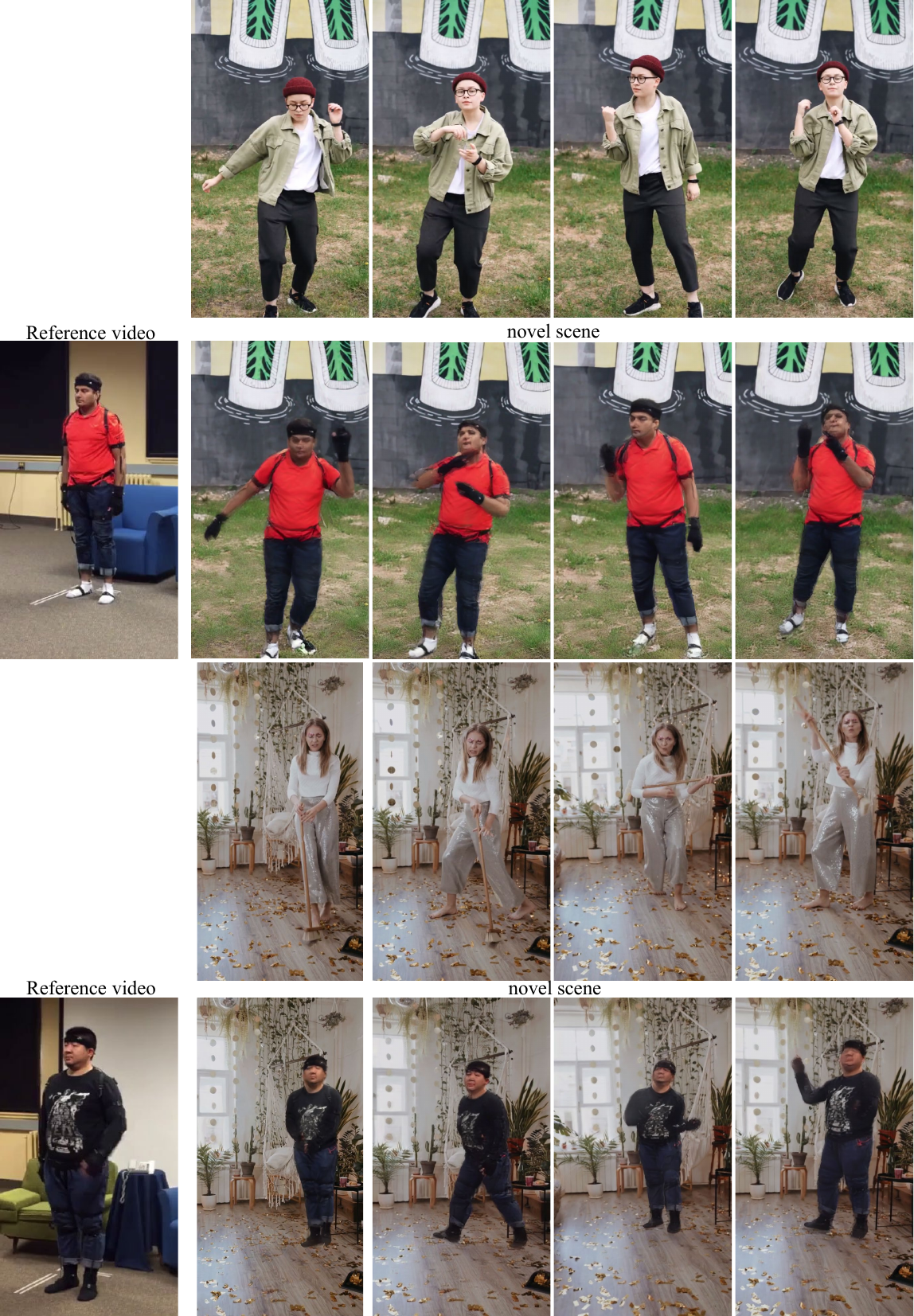} 
  \end{subfigure}
  \caption{Video reenactment showcasing the teleportation of characters into designated scenes, guided by predefined motions while preserving visual consistency and realism.}
  \label{fig:suppl3}
\end{figure*}

\noindent\textbf{Novel 3D Pose.} We present more visualization results of novel 3D poses generated using the AMASS motion capture dataset~\cite{AMASS:ICCV:2019}. These poses demonstrate the capability to generate lifelike human motions with diverse characteristics. In particular, the selected motions involve large-scale movements such as running, jumping, and dancing. Our method consistently generates coherent and stable motions while maintaining the overall appearance of the human subject. As shown in Figure~\ref{fig:suppl1}.

Additionally, as illustrated in Figure~\ref{fig:suppl1}, novel 3D poses are synthesized using the text-to-pose approach, MoMask~\cite{guo2024momask}. Under a variety of text-generated 3D poses, our method consistently performs well, highlighting its potential for more diverse and engaging applications. The model can also autonomously produce videos in scenarios involving significant motion translation, such as "jumping up" or "walking forward," without requiring any reference videos.

\noindent\textbf{Video Reenactment.} We also illustrate the model’s capability to adapt human motions across different environments, highlighting smooth transitions between various scene layouts. These results demonstrate the ability of the model to seamlessly teleport characters into designated scenes while preserving the consistency of the human subject and movement. As shown in Figure~\ref{fig:suppl3}, our approach ensures realistic human-scene interaction, maintaining temporal coherence and visual quality. Furthermore, Figure~\ref{fig:cross_video} demonstrates cross-video consistency during video generation, showcasing the model's ability to maintain coherence across different scenes and timeframes.

\section{Limitations}\label{sec:Appendix Limitations}
Although our method yields promising results, it does have some limitations. As shown in Figure~\ref{fig:suppl1} and Figure~\ref{fig:suppl3}, while the character pose guidance and scene transitions are smooth and visually consistent, occasional artifacts can appear when handling highly complex or dynamic scenes. These issues stem from the challenge of capturing intricate dynamic scene geometry by COLMAP~\cite{Schonberger_Frahm_2016,Schönberger_Zheng_Frahm_Pollefeys_2016}. Furthermore, despite the success of our approach with monocular videos, challenges remain in accurately rendering detailed facial expressions and fine-grained hand movements, which are areas we plan to address in future work.

\end{document}